\documentclass[letterpaper, 10 pt, journal, twoside]{IEEEtran}
\usepackage{amsmath,amsfonts}
\usepackage{algorithmic}
\usepackage{array}
 
\usepackage{textcomp}
\usepackage{stfloats}
\usepackage{url}
\usepackage{verbatim}
\usepackage{graphicx}
\usepackage{colortbl}

\usepackage{graphicx}
\usepackage{subcaption}
\def\BibTeX{{\rm B\kern-.05em{\sc i\kern-.025em b}\kern-.08em
    T\kern-.1667em\lower.7ex\hbox{E}\kern-.125emX}}
\usepackage{balance}
\usepackage[pagebackref=true,breaklinks=true,letterpaper=true,colorlinks,bookmarks=false]{hyperref}

\usepackage{graphicx}
\usepackage{amsmath}
\usepackage{amssymb}
\usepackage{booktabs}
\usepackage{multirow}
\usepackage{makecell}
\usepackage{caption}
\usepackage{subcaption}

\usepackage{hyperref}

\usepackage{tikz}
\usepackage{comment}
\usepackage{amsmath,amssymb} 
\usepackage{color}
\usepackage{bbding}

\begin{document}
\title{GMF: General Multimodal Fusion Framework for Correspondence Outlier Rejection}
\author{Xiaoshui Huang$^{1+}$, Wentao Qu$^{2+}$, Yifan Zuo$^{2*}$, Yuming Fang$^2$, Xiaowei Zhao$^3$
	\thanks{Manuscript received: May, 29, 2021; Revised August, 02, 2022; Accepted October, 31, 2022.}
	\thanks{This paper was recommended for publication by Editor Cesar Cadena Lerma upon evaluation of the Associate Editor and Reviewers' comments.
		This work was supported by in part by Shanghai AI Laboratory.} 
	\thanks{$^{1}$Shanghai AI Laboratory,  Shanghai, China, $^{2}$Jiangxi University of Finance and Economics, Jiangxi, China, $^{3}$Sany, China.}
	\thanks{$^{+}$Equal contribution, $^*$Corresponding authors: Yifan Zuo (\href{mailto:kenny0410@126.com}{kenny0410@126.com})}
	\thanks{Digital Object Identifier (DOI): see top of this page.}
}

\markboth{Submitted to IEEE Robotics and Automation Letters, Preprint version.}
{Huang \MakeLowercase{\textit{et al.}}: GMFNet} 

\maketitle

\begin{abstract}
	Rejecting correspondence outliers enables to boost the correspondence quality, which is a critical step in achieving { high point cloud registration accuracy}. The current state-of-the-art correspondence outlier rejection methods only utilize the structure features of the correspondences. However, texture information is critical to reject the correspondence outliers in our human vision system.  In this paper, we propose \textbf{G}eneral \textbf{M}ultimodal \textbf{F}usion (GMF) to {  learn to} reject the correspondence outliers by leveraging both the structure and texture information. Specifically, two cross-attention-based fusion layers are proposed to fuse the texture information from paired images and structure information from point correspondences. Moreover, we propose a convolutional position encoding layer to enhance the difference between $Tokens$ and enable the encoding feature pay attention to neighbor information. Our position encoding layer will make the cross-attention operation integrate both local and global information. Experiments on multiple datasets(3DMatch, 3DLoMatch, KITTI) and recent state-of-the-art models (3DRegNet, DGR, PointDSC) prove that our GMF achieves wide generalization ability and consistently improves the {  point cloud registration} accuracy. Furthermore, several ablation studies demonstrate the robustness of the proposed GMF on different loss functions, lighting conditions and noises. The code is available at \url{https://github.com/XiaoshuiHuang/GMF}.
	
\end{abstract}

\section{Introduction}
Point cloud registration is the cornerstone technology for numerous fields such as robotics and VR/AR. The current correspondence-based registration algorithms \cite{bai2021pointdsc,choy2020deep,pais20203dregnet,huang2021comprehensive} achieve the state-of-the-art registration accuracy in real-world point cloud datasets. Among the correspondence-based registration algorithms, accurate correspondences are the key to the registration accuracy. However, the repeatable and ambiguous structure patterns in real-world point clouds make the structure-based correspondences contain large outliers \cite{huang2021comprehensive}. These outliers will largely decrease the registration accuracy \cite{bai2021pointdsc}. This paper proposes a novel correspondence outlier rejection method to improve the registration accuracy.

\begin{figure}[t]
	\includegraphics[width=\linewidth]{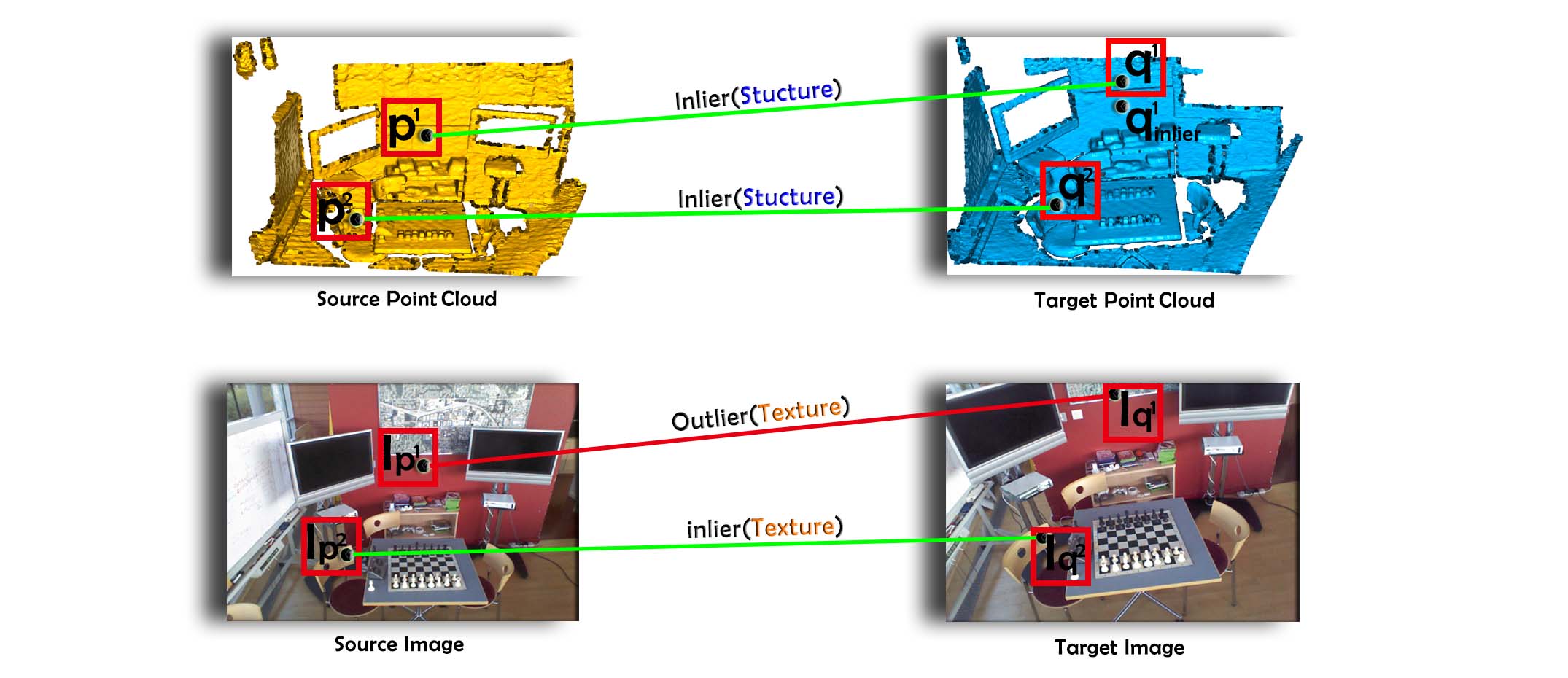}
	\caption{$p^1$ and $q^1$, $p^2$ and $q^2$ are two inliers based on structure information. However, $p^1$ and $q^1$ are actually a outlier ($p^1$ should correspond to $q^1_{inlier}$). The constraint of texture information can help correct the misclassification of structure information (The wall background of $p^1$ is red, $I_{p^1}$, however, the wall background of $q^1$ is white, $I_{q^1}$). }
	\label{f1}
\end{figure}
{ Correspondence outliers are the wrong correspondence.} Regarding the correspondence outlier rejection, there are mainly two kinds of algorithms: conventional methods \cite{aiger20084,fischler1981random,mellado2014super,zhou2016fast,yang2020teaser,huang2016coarse,huang2019fast} and learning-based methods \cite{bai2021pointdsc,choy2020deep,pais20203dregnet,huang2022unsupervised,huang2022robust,9919364}. Conventional methods utilize sample consensus strategy or additional constraints to reject the outliers { (wrong correspondences)}, such as RANSAC \cite{fischler1981random} and graph matching theory \cite{yang2020teaser}. These methods usually contain large search space to find the inliers. The limitation of traditional methods usually require large computational cost and the robustness faces a challenge to guarantee due to its large search space or approximation strategy. In comparison, learning-based methods find the initial correspondences first and feed the correspondences into a neural network for correspondence feature extraction. By this way, the correspondence outlier rejection problem is transformed into an outlier/inlier two-class classification problem. Typical examples of recent learning-based algorithms are 3DRegNet \cite{pais20203dregnet}, DGR \cite{choy2020deep}, PointDSC \cite{bai2021pointdsc}. However, the existing learning-based methods only utilize the structure information of correspondences while the structure only information faces challenges to reject the outliers in the repeatable and ambiguous structure patterns. Recently, there are several multi-modal registration methods \cite{9919364,park2017colored}. However, these methods focus on descriptors or optimization strategies.

This paper proposes a new general multimodal fusion (GMF) framework to reject the correspondence outliers by fusing both the texture and structure information. The principle is that the correspondence outliers could be further recognized by comparing their texture information. For example, Figure \ref{f1} shows an example of texture information shows the potential in rejecting the correspondence outlier ($p^1$ and $q^1$) that their structure information shows they are a correspondence but not in their texture information.   The proposed framework aims to explore whether the texture information is useful and how to use the texture information to reject the incorrect correspondences if it is useful. {  Our proposed framework is a general framework that utilizes the texture information for any based-learning outlier rejection methods. It is independent of the overall architecture and loss of the methods applied.}

Specifically, our proposed GMF contains two fusion layers and a local convolution position encoding (LCPE) layer. Inspired by recent success of Transformer \cite{xu2021co}, each fusion layer contains two MLPs layers and a cross-attention operation. The first fusion layer is utilized to fuse the paired image information and the second fusion layer aims to fuse image pair and point correspondence information. Because the cross-attention operation only integrates the global information, we propose a LCPE layer that aims to integrate local information for image pair and point correspondence. The benefits of our LCPE layer are two points. Firstly, it improves the discriminativeness of \emph{tokens} in the fusion layer. Secondly, it makes \emph{tokens} pay attention to other neighbor \emph{tokens}, which compensates for the cross-attention operation that only focuses on the global information. In summary, the contributions of this paper are listed as follows:
\begin{itemize}
	\item A general multimodal fusion module is proposed to reject the correspondence outliers by fusing structure and texture information.  
	
	\item A convolutional position encoding (LCPE) layer is proposed to integrate local information of neighbor tokens, which compensates for the cross-attention operation that only focuses on the global information.
	
	\item Comprehensive experiments on multiple datasets and models demonstrate the state-of-the-art (SOTA) accuracy, generalization and robustness of our method.
	
\end{itemize}

\section{Related Work}
Because the proposed algorithm aims to reject the correspondence outliers, this section reviews the related works about correspondence outlier rejection in the point cloud registration field. We review the related works from two aspects: conventional methods and learning-based methods.

\subsection{Conventional methods}
The conventional methods \cite{fischler1981random,aiger20084,zhou2016fast,mellado2014super,yang2020teaser,huang2016real} utilize optimization strategies to reject the correspondence outliers. These methods can further divided into two subcategories: sample consensus and constraint-based algorithms.

Sample consensus methods, such as RANSAC \cite{fischler1981random}, SDRSAC \cite{le2019sdrsac} and 4PCS \cite{aiger20084}, pre-define a rule to iteratively find a best transformation matrix and reject the outliers. The efficiency of these sample consensus methods will decrease dramatically when correspondence outlier percentage gets high. Super4PCS \cite{mellado2014super} uses a smart index to improve the efficiency of 4PCS algorithm and solves the point cloud registration in a linear time. However, the main limitation is that the sample consensus strategy relies on small subsets of the data to generate the hypotheses, e.g., the 4PCS considers only 4 congruent sets at a time. This is sub-optimal, as low overlapped point cloud pairs will contain a large percentage of outliers, thus making most of the hypotheses useless.

Constraint-based methods use additional constraint to reject outliers. Typical examples are FGR \cite{zhou2016fast} and TEASER \cite{yang2020teaser}. FGR \cite{zhou2016fast} added line constraint into the objective function for correspondence outlier rejection and formulated the point cloud registration problem into an optimization of Geman-McClure cost function. Despite its efficiency, FGR tends to fail when the outlier ratio is large but this is widely existed in low overlapped point clouds.  TEASER \cite{yang2020teaser} introduces pair-to-pair correspondence constraint and uses the graph maximal cliques theory for outlier rejection. \cite{enqvist2009optimal} introduces the point pair constraint into the graph-based objective function and rejects the outliers by solving the vertex cover problem. GORE \cite{bustos2017guaranteed} proposes a method to find the maximum consensus set to reject the outliers. This method is a preprocessing step to detect the inliers and then passes them to the following transformation estimation. CSGM \cite{huang2017systematic} uses graph matching theory to consider the neighbor correspondence consensus to reject the outliers and solve the point cloud registration.

However, all the above methods rely on the handcraft structure features and omit the texture information. Their performance still face a challenge in the repeatable and ambiguous structure patterns which are widely existed in the real-world point clouds.

\subsection{Learning-based methods}
The learning-based methods utilize a neural network to extract the feature of a correspondence and classify the correspondence as an outlier or inlier based on this feature. The current methods are mainly converted the outlier rejection problem into an outlier/inlier two-class classification problem.

DeepVCP \cite{lu2019deepvcp} proposes an end-to-end methods by integrating an outlier rejection module into the framework. 3DRegNet \cite{pais20203dregnet} utilizes fully connected layers and resNet-based neural network to extract feature for each correspondence, which is similar to pointNet. Then, the feature fed into a classification neural network to justify whether the correspondence is outlier. A parallel work, DGR \cite{choy2020deep} integrates the feature extraction, outlier rejection and transformation modules into an end-to-end framework. Specifically, DGR utilizes a sparse tensor convolution neural network to extract features for correspondences. Then, the features are utilized to regress inlier weights, e.g., 0.8 means the correspondence is 80\% confidence to be an inlier. These weights are combined with a weighted SVD to estimate the final transformation matrix. Following this end-to-end research pipeline, PointDSC \cite{bai2021pointdsc} designs a outlier rejection neural network by considering the spatial consistency constraint. Specifically, a neural network is utilized to extract line features for neighbor relations among correspondences. Then, these line features are propagated into each point correspondence for further inlier weight learning. These weights are utilized in a weighed SVD to estimate the transformation matrix. Similarly, \cite{shao2022scrnet} first calculated the average distance coding and average angle coding from a center point to the neighbor in a sphere cluster. Then, these codings feed into a MLP layer to extract a deep feature for outlier rejection. By this way, this paper also consider spatial consistency constraint into the neural network to detect the outliers. Recently, CofiNet \cite{yu2021cofinet} proposed an end-to-end coarse-to-fine algorithm to detect the inliers and reject the outliers based on the structure information.

However, all the above learning-based utilize the structure information only. They face the problem in the low overlapped point clouds because of the large outliers and ambiguous structures. This paper aims to solve the limitation of these problem by proposing a general multimodal fusion module that integrating texture information into the existing learning-based methods.

\section{The proposed general multimodal fusion (GMF)}

The proposed general multimodal fusion framework aims to leverage both texture and structure information to reject the correspondence outliers. To achieve this goal, we design three new neural network layers, two fusion layers and local convolutional position encoding(LCPE) layer. {   Figure \ref{f2} shows the overall framework. The input of top structure branch is point correspondences, and the input of bottom image branch is two images. 

\begin{figure*}[ht]
	\centering
	\includegraphics[width=0.9\linewidth]{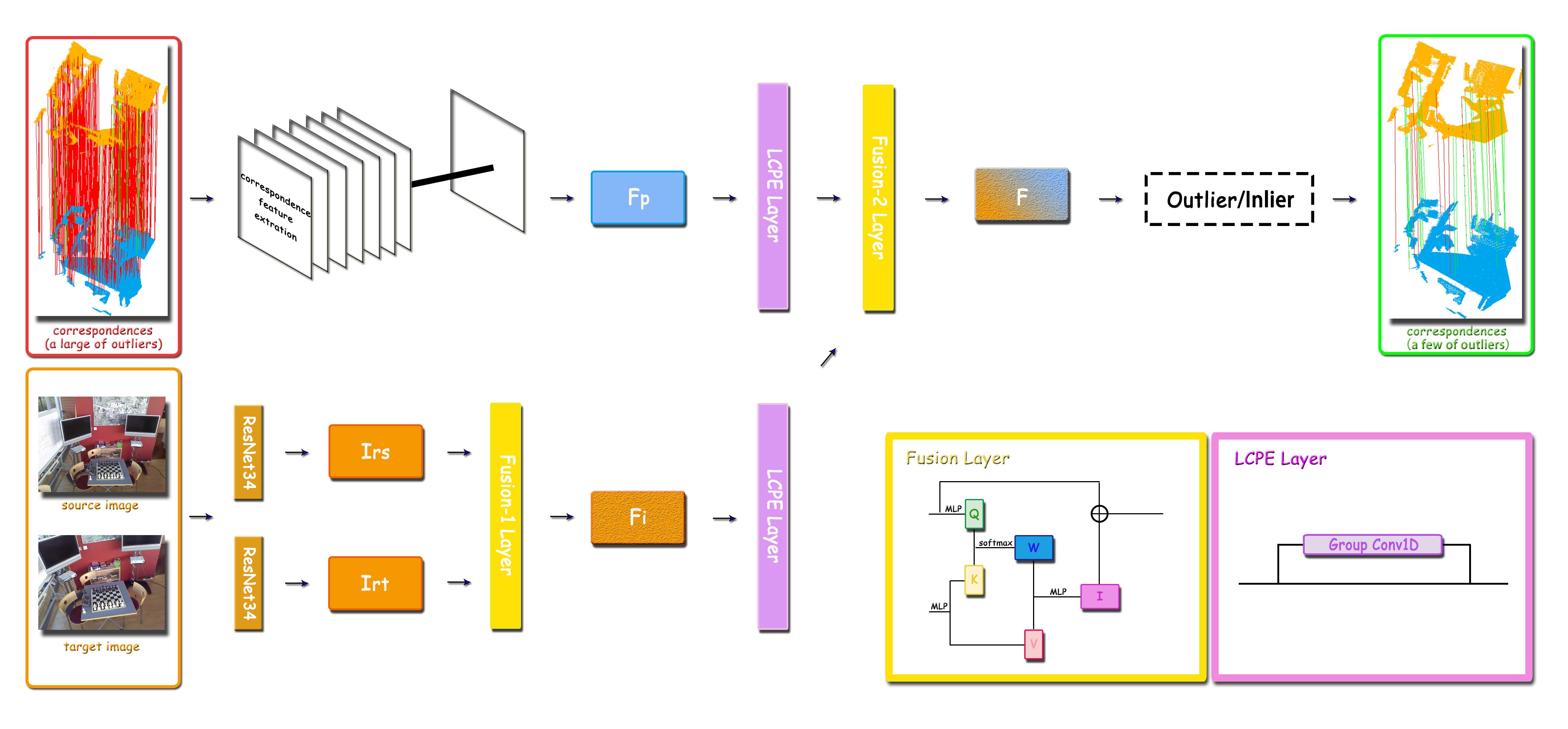}
	\caption{The overall of general multimodal fusion (GMF) framework. The $F_p$ is structure feature for point correspondence that can be extracted by many recent neural networks (e.g. DGR, PointDSC). $F_i$ is the fused texture feature by Fusion-1 layer. Then, structure feature $F_p$ and texture feature $F_i$ go through a local convolutional position encoding layer and then pass to Fusion-2 layer to get the final correspondence feature. This feature is then utilized to classify as outlier/inlier.}
	\label{f2}
\end{figure*}

 The fusion layers utilize the cross-attention mechanism to fuse the texture-texture (image branch) or texture-structure information. The LCPE layer utilizes the convolution operation to make $Tokens$ (pixel/point) pay attention to the local neighboring tokens, which makes the position encoding operation of our fusion layer contain both global and local $Tokens$ information.}

\subsection{Local Convolutional Position Encoding (LCPE)}
\label{lcpe}

{  Our objective is to integrate the local neighbor Token information for the attention operation. Therefore, the following cross-attention operation can extract features with both global and local information.}

Specifically, due to the unordered and unstructured property of point clouds, we consider each feature channel as a group and use the 1D group convolution to embed the local neighbor information on each channel space,  which could integrate local information for the following cross-attention operation. And it requires less resource consumption than conventional convolution.
Since the original features of paired images and point correspondences are from different domains, two LCPE are designed to add the position encoding for these features respectively.
\begin{eqnarray}
	\left\{
	\begin{aligned}
		F_1' = GC(F_1)_1+F_1 \\
		F_2' = GC(F_2)_2+F_2 \\
	\end{aligned}
	\right.
	\\
	GC(F) = CAT_{i=1}^C(w_i \cdot g_i)
	\label{e6}
\end{eqnarray}\\
where $F_1$ and $F_2$ represent features of different domains, respectively, and $GC(\cdot)$ represents 1D group convolution, and $F_1'$ and $F_2'$ are the output features with position encoding information.  $C$ is the feature dimension, $w_i$ is the convolution kernel, $g_i$ is the feature of each group, and $CAT(\cdot)$ is concatenating operation. In this paper, $F_1$ and $F_2$ represent the fused texture feature $F_i$ and correspondence feature $F_p$, respectively.  

\subsection{Fusion layers}
There are two fusion layers, one is to fuse the paired images, and the other is to fuse the texture and structure information. The main structure of these two fusion layers is a cross-attention module. Figure \ref{f2} visually shows the main structure.

\textbf{The first fusion layer (Fusion-1)} aims to integrate texture information from two images. 
We notice that the point correspondence feature contains the structure information of both the source and target point clouds, thus the texture features should also contain the information of both corresponding images. The \emph{Fusion-1} fuses the texture information of source image and target image by cross-attention operation to get the feature with richer texture information. 

Specifically, we first select the image paired that are corresponding to the source point cloud and target point cloud. Secondly, the source image $I_{s} \in R^{H \times W \times 3}$ and the target image $I_{t} \in R^{H \times W \times 3}$ are fed to the pre-trained ResNet34 to extract the image features $I_{rs} \in R^{H/8 \times W/8 \times C_i}$ and  $I_{rt} \in R^{H/8 \times W/8 \times C_i}$ . Thirdly, $I_{rs}$ is considered as $K_i \in R^{M_i \times C_t}$, $V_i \in R^{M_i \times C_t}$ and $I_{rt}$ is considered as $Q_i \in R^{M_i \times C_i}$($M_i = H/8*W/8$). Then, $K_i$, $V_i$ and $Q_i$ are utilized to obtain the texture feature $F_{st} \in R^{M_i \times C_i}$ that combines the texture information of the source point cloud image and the target point cloud image, 
\begin{eqnarray}
	F_{st} =MLP(softmax(\frac{Q_iK_i^T}{\sqrt{C_t}})*V_i)
	\label{e1}
\end{eqnarray}
where $C_t$ represents the intermediate dimension in cross-attention operation and is equal to $\frac{C_i}{2}$.

Finally, $F_{st}$ is processed by $GeLU$, and the features are further enhanced by residual connections to product the feature $F_{i} \in R^{M_i \times C_i}$ richer texture information,
\begin{eqnarray}
	F_{i} =MLP(GeLU(F_{st}) \cdot F_{st}) + I_{t}
	\label{e2}
\end{eqnarray}

\textbf{The second fusion layer (Fusion-2)} aims to fuse the structure information of point correspondences $F_p \in R^{M_p \times C_p}$ and the above fused texture information of paired images $F_i \in R^{M_i \times C_i}$. $F_p$ can be a correspondence feature of any dimension. 

Inspired by the recent Transformer \cite{xu2021co} in 2D images, we first use the above LCPE (introduced in Section \ref{lcpe}) to integrate the local neighbor $Tokens$ information in $F_i$ and $F_p$, respectively. Secondly, we use a layer normalization (LN) and a MLP to extract $K \in R^{M_i \times C_t}$, $V \in R^{M_i \times C_t}$ and $Q \in R^{M_p \times C_t}$, respectively.  Mathematically, the calculation of $K$, $V$, $Q$ can be presented as:
\begin{eqnarray}
	\left\{
	\begin{aligned}
		K = MLP(LN(LCPE_i(F_i)+F_i))\\
		V = MLP(LN(LCPE_i(F_i)+F_i))\\
		Q = MLP(LN(LCPE_p(F_p)+F_p))\\
	\end{aligned}
	\right.
	\label{e3}
\end{eqnarray}

Thirdly, we compute a weight matrix $W \in R^{M_p \times M_i}$ with a cross-attention operation $softmax(\frac{QK^T}{\sqrt{C_t}})$. $W$ represents the similarity  between pixels and point correspondences. The rational is that the $W$ will automatically select the texture to describe the correspondence. The feature with both texture and structure are more distinctive to the outliers/inliers classification problem.  

Fourthly, the texture information $I \in R^{M_p \times C_p}$ of point correspondences can be calculated by multiplying $W$ and $V$.
\begin{eqnarray}
	I =MLP(W \cdot V)
	\label{e4}
\end{eqnarray}

Finally, we obtain the fused feature $F \in R^{M_p \times C_p}$  that integrates both the structure and texture information as:
\begin{eqnarray}
	F =MLP(GeLU(I) \cdot I) + F_p
	\label{e5}
\end{eqnarray}
The $F$ describes the point correspondences and can be utilized to update the previous structure feature for correspondence outlier rejection.

\subsection{Loss $\&$ Training}
\textbf{Loss.} Because our GMF is a flexible framework that can combine with any correspondence feature extraction network to improve the previous feature by integrating the texture information, the loss functions could be exactly the same as the previous methods. Our GMF can improve the accuracy of existing methods at their loss functions. This shows that our GMF has a wide generalization ability without the restriction of loss functions.

\textbf{Training.} Similar to the generalization ability in loss functions, the training of our GMF depends on the parameter settings of the applied model itself. Our GMF does not need to ad-hoc design for the training strategies.  

\section{Experiments}
In this section, comprehensive experiments are performed to demonstrate the generalization performance and robustness of our GMF. Firstly, to demonstrate the wide generalization ability, we integrate our GMF framework with three recent end-to-end models (3DRegNet \cite{pais20203dregnet}, DGR \cite{choy2020deep}, PointDSC \cite{bai2021pointdsc}) and two descriptors(FPFH \cite{rusu2009fast}, FCGF \cite{choy2019fully}), and test them on both indoor and outdoor datasets (3DMatch \cite{zeng20173dmatch}, 3DLoMatch \cite{huang2021predator}, Kitti \cite{geiger2012we}).  Secondly, to demonstrate the robustness of our GMF, following \cite{wu2015deep}, we add even/uneven lighting conditions and random, salt, Gaussian noise on images to verify the robustness of our GMF. Finally, several ablation studies are conducted to verify the effectiveness of each module of our GMF.

\subsection{Experimental settings}
\textbf{DGR \cite{choy2020deep} \& PointDSC\cite{bai2021pointdsc}.} We follow the original training and evaluate processes of DGR and PointDSC. For the experiments of our method,  the GMF module is added to update the correspondence feature in DGR and pointDSC.  Then, the original correspondence feature is replaced with the updated feature. All the other settings are the same as the original algorithms to demonstrate the generalization ability of the proposed GMF. More details about how to add the GMF are described in the supplementary material.

\textbf{3DRegNet \cite{pais20203dregnet}.} It has been stated in PointDSC that 3DRegNet is difficult to converge on 3DMatch, thus we follow PointDSC to train the 3DRegNet on 3DMatch and 3DLoMatch datasets. The detail of implementation is described in the supplementary material.

\textbf{W/Wo.} 
In the experiment tables, {  ”W” means the original outlier rejection methods (e.g., PointDSC, 3DRegNet) with our GMF module (two fusion layers and LCPE layers) and ”Wo” means only the original methods without our GMF module.}

\subsection{Datasets and metrics}
\textbf{3DMatch \cite{zeng20173dmatch}.} 3DMatch is a widely used indoor point cloud dataset to evaluate the registration algorithms, which is captured by the RGBD sensor. The overlap ratios of point cloud pairs are larger than 30\% in this dataset. Following the experimental setting of PointDSC \cite{bai2021pointdsc}, we train all the models on the training data and evaluate them on the testing data.

\textbf{3DLoMatch \cite{huang2021predator}.} 3DLoMatch is a further development of 3DMatch by considering point clouds with an overlap ratio of 10\%-30\%. Following the Predator \cite{huang2021predator}, we train all the models on the training data and evaluate them on the testing data.

\textbf{KITTI \cite{geiger2012we}.} KITTI is a well-known outdoor dataset to evaluate the point cloud registration algorithms, which is captured by a 3D LiDAR sensor. Following the PointDSC \cite{bai2021pointdsc}, we train all the models and compare their accuracy. 

{  \textbf{Evaluation Metrics.} We use five metrics to evaluate our GMF: (1) Registration Recall(RR), the fraction of point cloud pairs that satisfy the accuracy threshold. (2) Rotation Error(RE), the mean rotation angle error. (3) Translation Error(TE), the mean translation error. (4) F1-score(F1), $F1 = \frac{2TP}{2TP+FN+FP}$. F1-score is used to measure the stability between precision and recall. (5) Inlier Recall(IR), the fraction of estimated correspondences whose residuals are below a certain threshold (i.e., 0.1m) under the ground-truth transformation. The first three metrics aim to evaluate the registration accuracy and the rest aims to evaluate the outlier rejection ability.}

\begin{figure*}[ht]
	\centering
	\includegraphics[width=0.8\linewidth,height=4.5cm ]{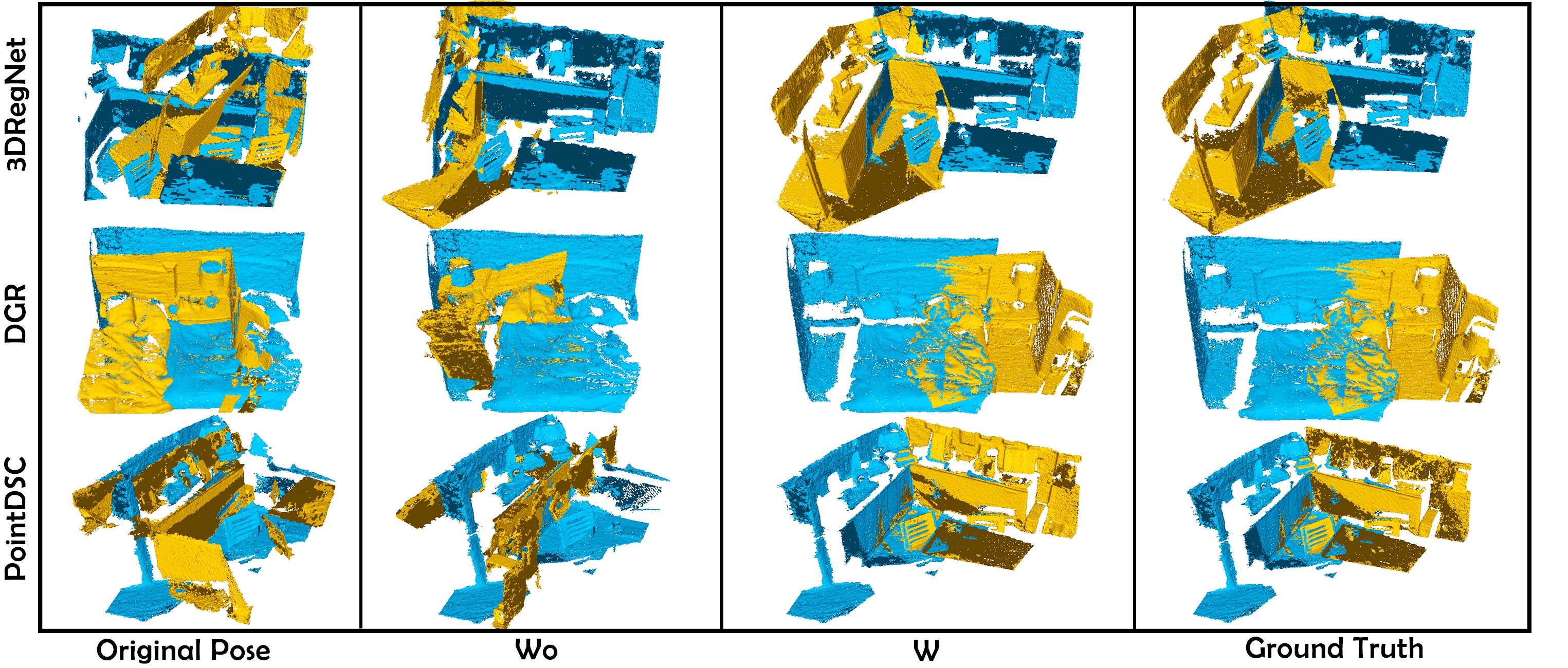}
	\caption{The visual comparison of generalization performance on 3DMatch.}
	\label{f3dm}
\end{figure*}
\subsection{Generalization Performance} 
In this section, we demonstrate the generalization performance by combining with three recent state-of-the-art methods (3DRegNet, DGR, PointDSC) and test them on three datasets (3DMatch, 3DLoMatch, KITTI). 

\subsubsection{Generalization on 3DMatch}
\begin{figure}[ht]
	\includegraphics[width=\linewidth]{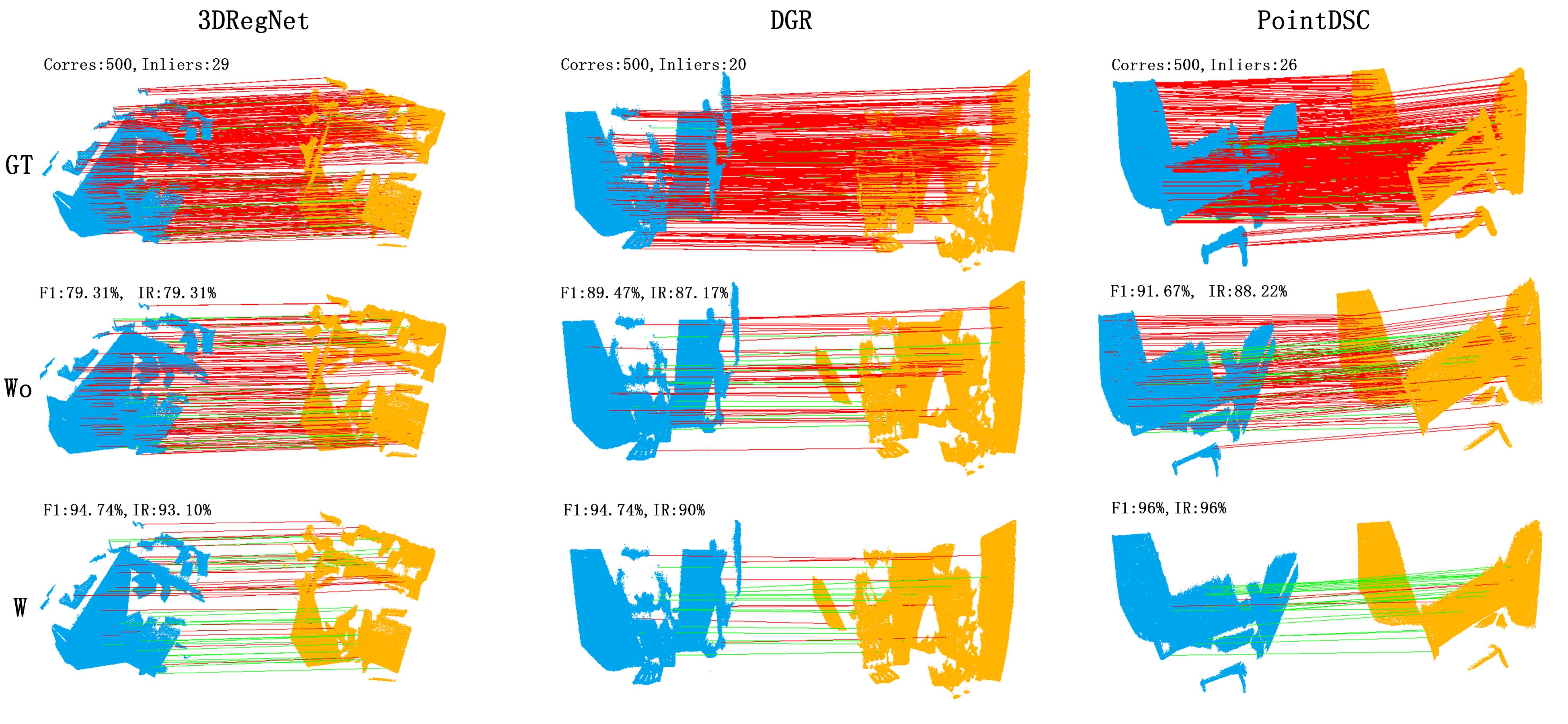}
	\caption{The visual comparison of correspondence outlier rejection on 3DMatch.}
	\label{fcorre}
\end{figure}

Table \ref{t3dm} shows the generalization performance of with/without our module on 3DMatch. We can see the registration recall (RR) is consistently improved on 3DMatch about 0.25\% $\sim$ 1.98\%, which demonstrates that our module can benefit the indoor point cloud registration. Figure \ref{f3dm} visually shows the better registration accuracy. In addition, we find that the proposed module can consistently improve the inlier recall (IR) by about 0.15\% $\sim$ 23.14\%. These experiments show that fusion of the texture information can enhance the distinctiveness of the correspondence features and benefit the correspondence outlier rejection on indoor point clouds. Figure \ref{fcorre} visually shows the better correspondence outlier rejection performance.

To further demonstrate the registration and correspondence outlier rejection performance on hand-craft point feature, following the PointDSC \cite{bai2021pointdsc}, we replace the deep learning point feature with FPFH \cite{rusu2009fast}. Table \ref{tfpfh} shows that our GMF framework can consistently improve the registration accuracy and inlier accuracy.

\begin{table}[h]

		\scriptsize
		\begin{tabular}{c|c|ccccc}	
			\hline
			\makecell[c]{\textbf{Method}} 
			&\makecell[c]{\textbf{W/Wo}}
			&{\textbf{RR}$(\%)$}    
			&{\textbf{RE}$(^\circ)$}
			&{\textbf{TE}$(cm)$} 
			&{\textbf{F1}$(\%)$}  
			&{\textbf{IR}$(\%$} \\
			
			\hline
			\makecell[c]{\multirow{2}{*}{3DRegNet}}
			&\makecell[c]{Wo}
			&\makecell[c]{71.41}                 &\makecell[c]{2.96}
			&\makecell[c]{8.69}                  &\makecell[c]{33.91}
			&\makecell[c]{29.70}        \\ 
			
			&\makecell[c]{W}
			&\makecell[c]{\textbf{72.95}}        &\makecell[c]{\textbf{1.99}}
			&\makecell[c]{\textbf{6.26}}         &\makecell[c]{\textbf{52.99}}
			&\makecell[c]{\textbf{52.84}}                 \\
			\hline
			
			\makecell[c]{\multirow{2}{*}{DGR}}
			&\makecell[c]{Wo}
			&\makecell[c]{85.20}                 &\makecell[c]{4.38}
			&\makecell[c]{17.23}                  &\makecell[c]{80.36}
			&\makecell[c]{73.58}                      \\             
			
			&\makecell[c]{W}
			&\makecell[c]{\textbf{87.18}}        &\makecell[c]{\textbf{3.20}}
			&\makecell[c]{\textbf{12.06}}        &\makecell[c]{\textbf{81.06}}
			&\makecell[c]{\textbf{77.47}}             \\
			\hline
			\makecell[c]{\multirow{2}{*}{DGR(s.g)}}
			&\makecell[c]{Wo}
			&\makecell[c]{91.30}                 &\makecell[c]{2.33}
			&\makecell[c]{8.56}                  &\makecell[c]{45.56}
			&\makecell[c]{82.24}                 \\             
			
			&\makecell[c]{W}
			&\makecell[c]{\textbf{93.28}}        &\makecell[c]{\textbf{2.04}}
			&\makecell[c]{\textbf{7.31}}         &\makecell[c]{\textbf{45.96}}
			&\makecell[c]{\textbf{83.01}}        \\
			\hline
			
			\makecell[c]{\multirow{2}{*}{PointDSC}}
			&\makecell[c]{Wo}
			&\makecell[c]{93.28}                 &\makecell[c]{\textbf{2.10}}
			&\makecell[c]{\textbf{6.52}}         &\makecell[c]{82.26}
			&\makecell[c]{86.41}                 \\
			
			&\makecell[c]{W}
			&\makecell[c]{\textbf{93.53}}        &\makecell[c]{2.16}
			&\makecell[c]{6.57}                  &\makecell[c]{\textbf{82.38}}
			&\makecell[c]{\textbf{86.56}}        \\
			\hline
			
		\end{tabular}

	\caption{Generalization performance evaluation on 3DMatch.}
	\label{t3dm}
\end{table}

\begin{table}[h]
	\begin{center}
		\scriptsize
		\begin{tabular}{c|c|ccccc}	
			\hline
			\makecell{\textbf{Method}} 
			&\makecell{\textbf{W/Wo}}
			&\makecell[c]{\textbf{RR}$(\%$}    &\makecell[c]{\textbf{RE}$(^\circ)$}
			&\makecell[c]{\textbf{TE}$(cm)$} 
			&\makecell[c]{\textbf{F1}$(\%$}  
			&\makecell[c]{\textbf{IR}$(\%)$} \\
			
			\hline
			\makecell[c]{\multirow{2}{*}{3DRegNet}}
			&\makecell[c]{Wo}
			&\makecell[c]{30.01}                 &\makecell[c]{\textbf{2.07}}
			&\makecell[c]{\textbf{6.15}}         &\makecell[c]{22.07}
			&\makecell[c]{21.45}               \\ 
			
			&\makecell[c]{W}
			&\makecell[c]{\textbf{32.72}}        &\makecell[c]{2.13}
			&\makecell[c]{6.33}                  &\makecell[c]{\textbf{23.04}}
			&\makecell[c]{\textbf{25.56}}        \\               
			\hline
			
			\makecell[c]{\multirow{2}{*}{DGR}}
			&\makecell[c]{Wo}
			&\makecell[c]{42.45}                 &\makecell[c]{13.78}
			&\makecell[c]{52.04}                  &\makecell[c]{12.13}
			&\makecell[c]{10.80}                      \\             
			
			&\makecell[c]{W}
			&\makecell[c]{\textbf{45.44}}        &\makecell[c]{\textbf{13.63}}
			&\makecell[c]{\textbf{51.71}}        &\makecell[c]{\textbf{37.65}}
			&\makecell[c]{\textbf{41.56}}             \\
			\hline	 
			
			\makecell[c]{\multirow{2}{*}{DGR(s.g)}}
			&\makecell[c]{Wo}
			&\makecell[c]{69.13}                 &\makecell[c]{5.90}
			&\makecell[c]{21.83}                  &\makecell[c]{17.35}
			&\makecell[c]{12.42}                 \\             
			
			&\makecell[c]{W}
			&\makecell[c]{\textbf{77.57}}        &\makecell[c]{\textbf{5.44}}
			&\makecell[c]{\textbf{21.65}}         &\makecell[c]{\textbf{67.00}}
			&\makecell[c]{\textbf{67.43}}        \\
			\hline
			
			\makecell[c]{\multirow{2}{*}{PointDSC}}
			&\makecell[c]{Wo}
			&\makecell[c]{93.28}                 &\makecell[c]{\textbf{2.10}}
			&\makecell[c]{\textbf{6.52}}         &\makecell[c]{69.85}
			&\makecell[c]{71.61}                 \\
			
			&\makecell[c]{W}
			&\makecell[c]{\textbf{93.53}}        &\makecell[c]{2.16}
			&\makecell[c]{6.57}                  &\makecell[c]{\textbf{73.26}}
			&\makecell[c]{\textbf{76.00}}        \\
			\hline
			
		\end{tabular}
	\end{center}
	\caption{Generalization performance evaluation on 3DMatch using FPFH.}
	\label{tfpfh}
\end{table}

\subsubsection{Generalization on 3DLoMatch}
\begin{figure}[ht]
	\includegraphics[width=\linewidth]{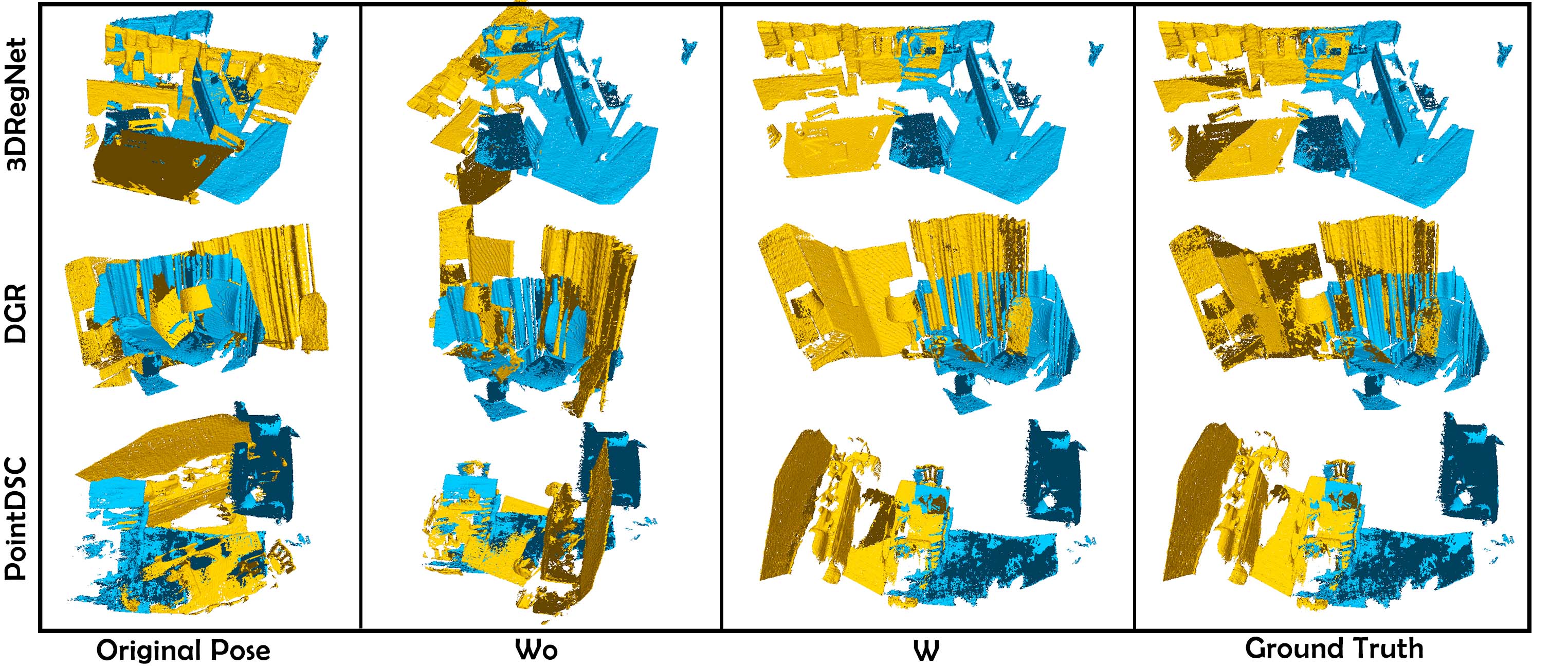}
	\caption{The visual comparison of generalization performance on 3DLoMatch. }
	\label{flomatch}
\end{figure}

Table \ref{t3dlm} shows the generalization performance on 3DLoMatch. These experiments show that our proposed framework can consistently improve the registration accuracy and correspondence outlier rejection performance on low-overlapped point clouds. The reason is that our fusion layers can extract the discriminative texture information to describe the correspondence so that benefits the correspondence outlier rejection. Figure \ref{flomatch} visually shows the our better registration accuracy on low-overlapped point clouds.

\begin{table}[h]
	\begin{center}
		\scriptsize
		\begin{tabular}{c|c|ccccc}	
			\hline
			\makecell{\textbf{Method}} 
			&\makecell{\textbf{W/Wo}}
			&\makecell[c]{\textbf{RR}$(\%)$}    &\makecell[c]{\textbf{RE}$(^\circ)$}
			&\makecell[c]{\textbf{TE}$(cm)$} 
			&\makecell[c]{\textbf{F1}$(\%)$}  
			&\makecell[c]{\textbf{IR}$(\%)$} \\
			
			\hline
			\makecell[c]{\multirow{2}{*}{3DRegNet}}
			&\makecell[c]{Wo}
			&\makecell[c]{11.85}                 &\makecell[c]{5.20}
			&\makecell[c]{15.16}                  &\makecell[c]{8.65}
			&\makecell[c]{8.30}                 \\ 
			
			&\makecell[c]{W}
			&\makecell[c]{\textbf{12.52}}        &\makecell[c]{\textbf{3.34}}
			&\makecell[c]{\textbf{11.46}}         &\makecell[c]{\textbf{10.79}}
			&\makecell[c]{\textbf{9.95}}        \\
			\hline
			
			\makecell[c]{\multirow{2}{*}{DGR}}
			&\makecell[c]{Wo}
			&\makecell[c]{44.35}                     &\makecell[c]{13.68}
			&\makecell[c]{52.19}                     &\makecell[c]{39.87}
			&\makecell[c]{36.47}                      \\             
			
			&\makecell[c]{W}
			&\makecell[c]{\textbf{45.04}}        &\makecell[c]{\textbf{14.25}}
			&\makecell[c]{\textbf{57.07}}        &\makecell[c]{\textbf{40.12}}
			&\makecell[c]{\textbf{39.27}}             \\
			\hline	 
			
			\makecell[c]{\multirow{2}{*}{DGR(s.g)}}
			&\makecell[c]{Wo}
			&\makecell[c]{54.60}                 &\makecell[c]{11.26}
			&\makecell[c]{41.76}                &\makecell[c]{45.46}
			&\makecell[c]{47.78}                 \\             
			
			&\makecell[c]{W}
			&\makecell[c]{\textbf{58.73}}        &\makecell[c]{\textbf{10.93}}
			&\makecell[c]{\textbf{40.76}}         &\makecell[c]{\textbf{45.96}}
			&\makecell[c]{\textbf{48.28}}        \\
			\hline
			
			\makecell[c]{\multirow{2}{*}{PointDSC}}
			&\makecell[c]{Wo}
			&\makecell[c]{56.32}                 &\makecell[c]{3.54}
			&\makecell[c]{11.53}                   &\makecell[c]{46.73}
			&\makecell[c]{51.58}                 \\
			
			&\makecell[c]{W}
			&\makecell[c]{\textbf{56.82}}        &\makecell[c]{\textbf{3.52}}
			&\makecell[c]{\textbf{11.50}}       &\makecell[c]{\textbf{47.65}}
			&\makecell[c]{\textbf{52.56}}        \\
			\hline
			
		\end{tabular}
	\end{center}
	\caption{Generalization performance evaluation on 3DLoMatch.}
	\label{t3dlm}
\end{table}

\begin{figure*}[ht]
	\centering
	\includegraphics[width=0.8\linewidth,height=4.5cm]{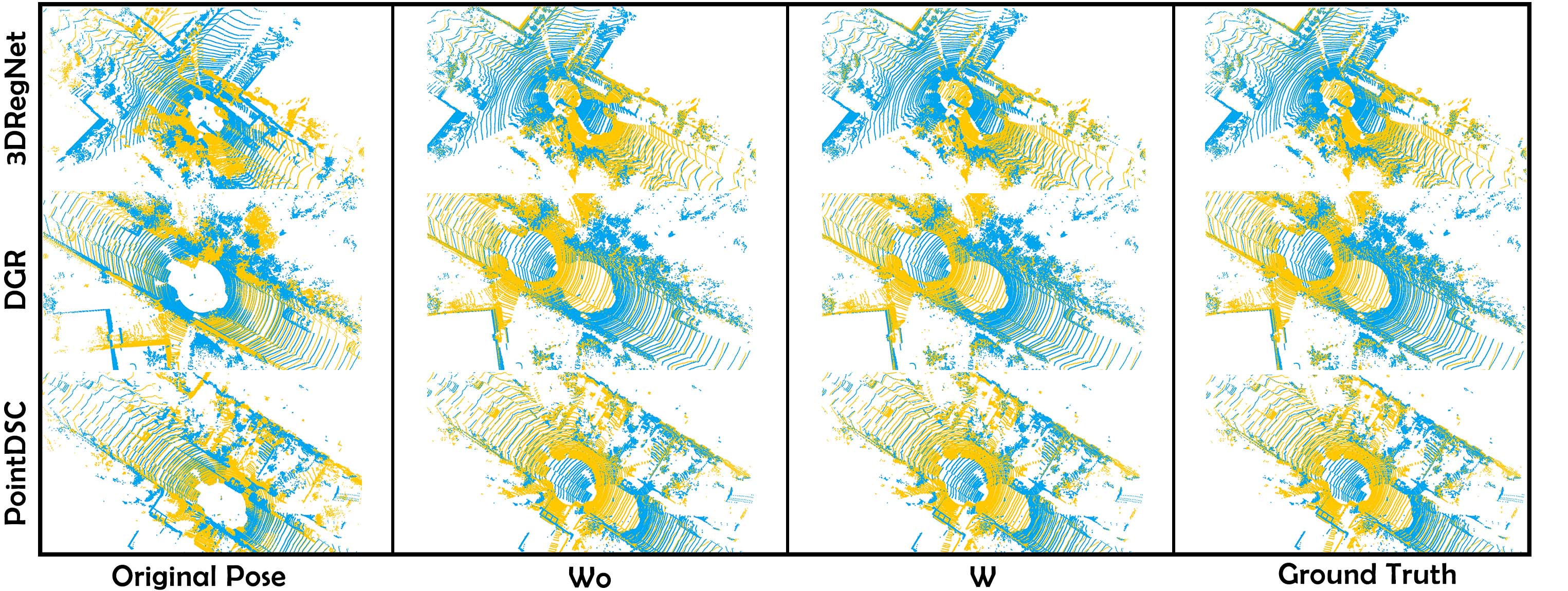}
	\caption{The visual comparison of generalization performance on KITTI.}
	\label{fkitti}
\end{figure*}

\subsubsection{Generalization on KITTI}
To further demonstrate the generalization performance, we also evaluate our module on outdoor dataset. Table \ref{tkitti} shows that the registration accuracy improve about 0.36\% $\sim$ 2.7\% and the inlier accuracy improve about 1.65\% $\sim$ 9.8\%. Particularly, our general multimodal fusion framework can improve the inlier accuracy of PointDSC by 9.8\%. The big performance improvement on the state-of-the-art method shows that the fusion of texture information is important for outdoor point cloud registration. Figure \ref{fkitti} visually shows the our better registration accuracy on outdoor point clouds.

\begin{table}[h]
	\begin{center}
		\scriptsize
		\begin{tabular}{c|c|ccccc}	
			\hline
			\makecell{\textbf{Method}} 
			&\makecell{\textbf{W/Wo}}
			&\makecell[c]{\textbf{RR}$(\%)$}    &\makecell[c]{\textbf{RE}$(^\circ)$}
			&\makecell[c]{\textbf{TE}$(cm)$} 
			&\makecell[c]{\textbf{F1}$(\%)$}  
			&\makecell[c]{\textbf{IR}$(\%)$} \\
			
			\hline
			\makecell[c]{\multirow{2}{*}{3DRegNet}}
			&\makecell[c]{Wo}
			&\makecell[c]{95.32}                 &\makecell[c]{0.52}
			&\makecell[c]{22.59}                 &\makecell[c]{71.74}  
			&\makecell[c]{85.09}\\ 
			
			&\makecell[c]{W}
			&\makecell[c]{\textbf{98.02}}        &\makecell[c]{\textbf{0.49}}
			&\makecell[c]{\textbf{21.04}}        &\makecell[c]{\textbf{85.09}}
			&\makecell[c]{\textbf{93.27}}  
			\\
			
			\hline
			
			\makecell[c]{\multirow{2}{*}{DGR}}
			&\makecell[c]{Wo}
			&\makecell[c]{98.20}                 &\makecell[c]{0.43}
			&\makecell[c]{23.28}           &\makecell[c]{96.82}      
			&\makecell[c]{73.60}\\             
			
			&\makecell[c]{W}
			&\makecell[c]{\textbf{98.73}}        &\makecell[c]{\textbf{0.73}}
			&\makecell[c]{\textbf{16.23}}         &\makecell[c]{\textbf{97.18}}     
			&\makecell[c]{\textbf{75.25}}\\
			\hline
			
			\makecell[c]{\multirow{2}{*}{PointDSC}}
			&\makecell[c]{Wo}
			&\makecell[c]{98.02}                 &\makecell[c]{\textbf{0.46}}
			&\makecell[c]{\textbf{21.05}}       &\makecell[c]{85.37}
			&\makecell[c]{81.48}                \\
			
			&\makecell[c]{W}
			&\makecell[c]{\textbf{98.38}}        &\makecell[c]{0.47}
			&\makecell[c]{21.09}                 &\makecell[c]{\textbf{86.93}}
			&\makecell[c]{\textbf{91.28}}\\
			\hline
			
		\end{tabular}
	\end{center}
	\caption{Generalization performance evaluation on KITTI.}
	\label{tkitti}
\end{table}

\subsection{Robustness analysis} 
In this section, we will conduct several experiments on 3DMatch to demonstrate the robustness of the proposed GMF. 

\subsubsection{Sensitivity on different losses.}
Firstly, we want to evaluate the sensitivity of our GMF to different losses in the model. We conduct this ablation study on the current best outlier rejection method PointDSC.

PointDSC has three losses: 1) BCE loss(BCE) 2) Spectral Matching loss(SM) 3) Transformation loss(T). We keep the BCE loss for the outlier/inlier classification and evaluate the loss sensitivity of GMF+PointDSC by keeping/removing the other two losses. Table \ref{t4} shows that our GMF achieves consistent improvement at different loss functions, which demonstrates that our GMF is completely unconstrained to the loss functions of the applied models. The reason is that our GMF plays the role to enhance the distinctiveness of correspondence features by fusing structure information and texture information, which does not affect the loss functions.

\begin{table}[h]
	\begin{center}
		\scriptsize
		\begin{tabular}{c|c|ccccc}	
			\hline
			\makecell{\textbf{Method}} 
			&\makecell{\textbf{W/Wo}}
			&\makecell[c]{\textbf{RR}$(\%)$}    &\makecell[c]{\textbf{RE}$(^\circ)$}
			&\makecell[c]{\textbf{TE}$(cm)$} 
			&\makecell[c]{\textbf{F1}$(\%)$}  
			&\makecell[c]{\textbf{IR}$(\%)$} \\
			\hline
			
			\makecell[c]{\multirow{2}{*}{BCE + SM}}
			&\makecell[c]{Wo}
			&\makecell[c]{92.61}                 &\makecell[c]{\textbf{2.10}}
			&\makecell[c]{\textbf{6.48}}         &\makecell[c]{81.78}
			&\makecell[c]{85.82} \\
			
			&\makecell[c]{W}
			&\makecell[c]{\textbf{93.41}}        &\makecell[c]{2.12}
			&\makecell[c]{6.56}                  &\makecell[c]{\textbf{82.50}}
			&\makecell[c]{\textbf{86.83}}        \\
			\hline
			
			\makecell[c]{\multirow{2}{*}{BCE + T}}
			&\makecell[c]{Wo}
			&\makecell[c]{92.79}                 &\makecell[c]{\textbf{2.08}}
			&\makecell[c]{\textbf{6.52}}         &\makecell[c]{81.93}
			&\makecell[c]{85.97}                 \\             
			
			&\makecell[c]{W}
			&\makecell[c]{\textbf{93.28}}        &\makecell[c]{2.12}
			&\makecell[c]{\textbf{6.52}}         &\makecell[c]{\textbf{82.44}}
			&\makecell[c]{\textbf{86.59}}        \\
			\hline
			\makecell[c]{\multirow{2}{*}{BCE}}
			&\makecell[c]{Wo}
			&\makecell[c]{92.42}                 &\makecell[c]{\textbf{2.07}}
			&\makecell[c]{\textbf{6.58}}         &\makecell[c]{79.74}
			&\makecell[c]{81.73}                 \\             
			
			&\makecell[c]{W}
			&\makecell[c]{\textbf{93.28}}        &\makecell[c]{2.13}
			&\makecell[c]{6.60}                  &\makecell[c]{\textbf{82.35}}
			&\makecell[c]{\textbf{86.62}}        \\
			\hline
			
		\end{tabular}
	\end{center}
	\caption{The performance of our GMF by removing different losses on PointDSC.}
	\label{t4}
\end{table}

\subsubsection{Robustness to lighting condition.} 
Since our GMF needs to select discriminative texture information from image features, we verify the robustness on various light conditions and image noises. 

Inspired by \cite{wu2015deep}, we add varying lighting conditions by adding an even or uneven brightness map to one of the corresponding images. Table \ref{t5} shows that the proposed method is robust to mildly varying lighting conditions. We can notice that there is still more than $1.61\%$ performance improvement for DGR under different lighting conditions. 

\begin{table}[h]
	\begin{center}
		\scriptsize
		\begin{tabular}{c|c|ccccc}	
			\hline
			\makecell{\textbf{Method}} 
			&\makecell{\textbf{Interval}}
			&\makecell[c]{\textbf{RR}$(\%)$}    &\makecell[c]{\textbf{RE}$(^\circ)$}
			&\makecell[c]{\textbf{TE}$(cm)$} 
			&\makecell[c]{\textbf{F1}$(\%)$}  
			&\makecell[c]{\textbf{IR}$(\%)$} \\
			\hline
			\makecell[c]{DGR}
			&\makecell[c]{-}
			&\makecell[c]{91.30}               &\makecell[c]{2.33}
			&\makecell[c]{8.56}                &\makecell[c]{80.36}
			&\makecell[c]{82.24}               \\
			
			\makecell[c]{+GMF}
			&\makecell[c]{-}
			&\makecell[c]{\textbf{93.22}}      &\makecell[c]{2.03}
			&\makecell[c]{7.47}                &\makecell[c]{81.06}
			&\makecell[c]{83.00}               \\
			
			\cline{1-2}
			
			\makecell[c]{\multirow{4}{*}{Even}}
			&\makecell[c]{$[0.3,1.8]$}
			&\makecell[c]{93.10}               &\makecell[c]{2.12}
			&\makecell[c]{7.68}                &\makecell[c]{80.94}
			&\makecell[c]{82.87}               \\
			
			&\makecell[c]{$[0.5,1.5]$}
			&\makecell[c]{93.16}               &\makecell[c]{2.05}
			&\makecell[c]{7.48}                &\makecell[c]{\textbf{81.12}}
			&\makecell[c]{\textbf{83.05}}      \\
			
			&\makecell[c]{$[0.75,1.25]$}
			&\makecell[c]{93.10}               &\makecell[c]{2.02}
			&\makecell[c]{7.48}                &\makecell[c]{81.03}
			&\makecell[c]{82.96}               \\           
			
			&\makecell[c]{$[0.9,1.1]$}
			&\makecell[c]{93.04}               &\makecell[c]{2.09}
			&\makecell[c]{7.86}                &\makecell[c]{80.96}
			&\makecell[c]{82.90}               \\
			\cline{1-2}
			\makecell[c]{\multirow{4}{*}{Uneven}}
			&\makecell[c]{$[0.3,1.8]$}
			&\makecell[c]{92.91}               &\makecell[c]{2.12}
			&\makecell[c]{7.92}                &\makecell[c]{80.90}
			&\makecell[c]{82.85}               \\
			
			&\makecell[c]{$[0.5,1.5]$}
			&\makecell[c]{93.10}               &\makecell[c]{\textbf{2.00}}
			&\makecell[c]{\textbf{7.45}}       &\makecell[c]{81.15}
			&\makecell[c]{83.09}               \\
			
			&\makecell[c]{$[0.75,1.25]$}
			&\makecell[c]{92.98}               &\makecell[c]{2.18}
			&\makecell[c]{8.08}                 &\makecell[c]{80.86}
			&\makecell[c]{82.77}              \\           
			
			&\makecell[c]{$[0.9,1.1]$}
			&\makecell[c]{92.98}               &\makecell[c]{2.06}
			&\makecell[c]{7.57}                &\makecell[c]{80.91}
			&\makecell[c]{82.84}               \\
			\hline
			
		\end{tabular}
	\end{center}
	\caption{The performance of our GMF under different lighting condition on DGR. The second column shows the different lighting interval. The larger the interval, the greater the light changes for the images.}
	\label{t5}
\end{table}

\subsubsection{Robustness to noise.} 
We also consider the robustness of our GMF to random noise, salt noise and Gaussian noise. Table \ref{t6} shows that our GMF is robust to various noises in image and more than $1.68\%$ performance improvement for DGR.  The reason is that the ICA model of our GMF learns a weight matrix to globally select the related texture information to describe the correspondences. The noise in the texture information has little impact to this global selection strategy.

\begin{table}[h]
	\begin{center}
		\scriptsize
		\begin{tabular}{c|ccccc}	
			\hline
			\makecell{\textbf{Noisy}} 
			&\makecell[c]{\textbf{RR}$(\%)$}    &\makecell[c]{\textbf{RE}$(^\circ)$}
			&\makecell[c]{\textbf{TE}$(cm)$} 
			&\makecell[c]{\textbf{F1}$(\%)$}  
			&\makecell[c]{\textbf{IR}$(\%)$} \\
			\hline
			\makecell[c]{DGR}
			&\makecell[c]{91.30}               &\makecell[c]{2.33}
			&\makecell[c]{8.56}                &\makecell[c]{80.36}
			&\makecell[c]{82.24}               \\
			
			\makecell[c]{+GMF}
			&\makecell[c]{\textbf{93.22}}      &\makecell[c]{\textbf{2.03}}
			&\makecell[c]{7.47}                &\makecell[c]{81.06}
			&\makecell[c]{83.00}               \\        
			
			\makecell[c]{Random Noise}
			&\makecell[c]{92.98}               &\makecell[c]{2.11}               
			&\makecell[c]{7.76}                &\makecell[c]{81.05}          
			&\makecell[c]{82.97}               \\ 
			
			\makecell[c]{Salt Noise}
			&\makecell[c]{92.98}               &\makecell[c]{2.06}               
			&\makecell[c]{\textbf{7.44}}       &\makecell[c]{81.01}          
			&\makecell[c]{82.92}               \\  
			
			\makecell[c]{Gaussian Noise}
			&\makecell[c]{93.10}               &\makecell[c]{2.08}               
			&\makecell[c]{7.53}                &\makecell[c]{\textbf{81.07}}         
			&\makecell[c]{\textbf{83.02}}      \\  
			
			\hline	 
		\end{tabular}
	\end{center}
	\caption{The performance of our GMF under different image noises on DGR.}
	\label{t6}
\end{table}

{ 
\subsubsection{Discussion of texture missing.} 
Our method is designed for multimodal data, because the current point cloud sensors are becoming consumer-affordable, and many vision system contain multiple types of vision sensors (Lidar and RGB camera). The point cloud and RGB image can be acquired at the same time by many vision systems.Sometimes, we may only capture point cloud, our method is also available by rendering an image from the point cloud. 
}

\subsection{Ablation study} 
In this section, we will conduct several ablation studies to demonstrate the effectiveness of the proposed GMF.

\textbf{Retrieve texture with camera parameters.} We use camera parameters project the point cloud on image to get RGB texture for each point. Then, we concatenate the texture features (extract by ResNet34) and structure features on channel dimension.  Table \ref{t7} show that the proposed GMF is better than the direct feature concatenation on channel dimension for multimodal fusion.


\begin{table}[h]
	\begin{center}
		\scriptsize
		\begin{tabular}{c|ccccc}	
		\hline
		      \makecell{\textbf{Method}} 
             &\makecell[c]{\textbf{RR}$(\%)$}    &\makecell[c]{\textbf{RE}$(^\circ)$}
             &\makecell[c]{\textbf{TE}$(cm)$}  
             &\makecell[c]{\textbf{IR}$(\%)$}  &\makecell[c]{\textbf{F1}$(\%)$}\\
        \hline

              \makecell[c]{Concat}
             &\makecell[c]{61.98}                 &\makecell[c]{2.79}
			 &\makecell[c]{8.46}                  &\makecell[c]{37.76}
			 &\makecell[c]{39.43}        \\ 

              \makecell[c]{+GMF}
             &\makecell[c]{\textbf{72.95}}        &\makecell[c]{\textbf{1.99}}
			 &\makecell[c]{\textbf{6.26}}         &\makecell[c]{\textbf{52.99}}
			 &\makecell[c]{\textbf{52.84}}\\ 			 

		\hline	 
		\end{tabular}
	\end{center}
	\caption{Ablation study of fusion methods on 3DRegNet.}
	\label{t7}
\end{table}

{ \textbf{Baseline.}
We designed a baseline on 3DRegNet to reject the wrong correspondence (outliers). Specifically, first, we get the correspondence by using FCGF. Then, we project the point coordinates of the correspondences onto the RGB image through the camera intrinsic parameter to get the RGB values for every 3d point. Next, we concatenate the RGB with the XYZ coordinates of the points on the channel dimension. Namely, the input correspondence feature is changed from [x1 y1, z1, x2, y2, z2] to [x1, y1, z1, r1, g1, b1, x2, y2, z2, r2, g2, b2]. Finally, the deep correspondence features are extracted by 3DRegNet. These deep correspondence features are used to reject the correspondence outliers. We also added our GMF to the baseline to see the performance gain.

The below Table \ref{baseline} shows that the performance of the baseline is worse than the performance of original 3DRegNet. 
The reason is that the camera parameters may contain flaws so that the projection between point cloud and image may not perfectly aligned. With the wrong point and pixel matches, the direct coordinate concatenation obtains worse results than the original 3DRegNet. However, our GMF can automatically select the most effective texture information to describe the correspondences by transformer’s global ability, which can avoid this problem.
}

\begin{table}[h]
	\begin{center}
		\scriptsize
		\begin{tabular}{c|ccccc}	
		\hline
		      \makecell{\textbf{Method}} 
             &\makecell[c]{\textbf{RR}$(\%)$}    &\makecell[c]{\textbf{RE}$(^\circ)$}
             &\makecell[c]{\textbf{TE}$(cm)$}  
             &\makecell[c]{\textbf{IR}$(\%)$}  &\makecell[c]{\textbf{F1}$(\%)$}\\
        \hline
              \makecell[c]{3DRegNet}
             &\makecell[c]{71.41}                 &\makecell[c]{2.96}
			 &\makecell[c]{8.69}                  &\makecell[c]{33.91}
			 &\makecell[c]{29.70}        \\ 
            
              \makecell[c]{Baseline}
             &\makecell[c]{70.80}        
             &\makecell[c]{2.60}
			 &\makecell[c]{7.88}         &\makecell[c]{25.61}
			 &\makecell[c]{28.84}\\ 
            
              \makecell[c]{+GMF}
             &\makecell[c]{\textbf{72.95}}        &\makecell[c]{\textbf{1.99}}
			 &\makecell[c]{\textbf{6.26}}         &\makecell[c]{\textbf{52.99}}
			 &\makecell[c]{\textbf{52.84}}\\ 			 

		\hline	 
		\end{tabular}
	\end{center}
	\caption{Ablation study of baseline on 3DRegNet.}
	\label{baseline}
\end{table}

\textbf{Fusion-1, Fusion-2, LCPE.} We verify the effectiveness of three modules (Fusion-1, Fusion-2, LCPE) of our GMF in the ablation study. We take 3DRegNet as an example to verify the effectiveness of each module by adding and removing different modules of GMF. Table \ref{t8} shows that our three modules can improve both registration accuracy and correspondence inlier accuracy. 

\begin{table}[h]
	\begin{center}
		\scriptsize
		\begin{tabular}{c|c|c|ccccc}	
		\hline
		      \makecell{\textbf{F-2}} 
		     &\makecell{\textbf{F-1}} 
		     &\makecell{\textbf{LCPE}}
             &\makecell[c]{\textbf{RR}$(\%)$}    &\makecell[c]{\textbf{RE}$(^\circ)$}
             &\makecell[c]{\textbf{TE}$(cm)$}  
             &\makecell[c]{\textbf{IR}$(\%)$}  &\makecell[c]{\textbf{F1}$(\%)$}\\
        \hline
              \makecell[c]{\XSolidBrush}
             &\makecell[c]{\XSolidBrush}
             &\makecell[c]{\XSolidBrush}               
             &\makecell[c]{71.41}                 &\makecell[c]{2.96}
			 &\makecell[c]{8.69}                  &\makecell[c]{33.91}
			 &\makecell[c]{29.70}        \\ 
			 
              \makecell[c]{\Checkmark}
             &\makecell[c]{\XSolidBrush}
             &\makecell[c]{\XSolidBrush}               
             &\makecell[c]{72.02}                 &\makecell[c]{2.74}
			 &\makecell[c]{7.13}                  &\makecell[c]{42.44}
			 &\makecell[c]{35.15}        \\ 

              \makecell[c]{\Checkmark}
             &\makecell[c]{\Checkmark}
             &\makecell[c]{\XSolidBrush}               
             &\makecell[c]{72.53}                 &\makecell[c]{2.25}
			 &\makecell[c]{6.54}                  &\makecell[c]{48.45}
			 &\makecell[c]{49.03}        \\ 

              \makecell[c]{\Checkmark}
             &\makecell[c]{\Checkmark}
             &\makecell[c]{\Checkmark}                 
             &\makecell[c]{\textbf{72.95}}        &\makecell[c]{\textbf{1.99}}
			 &\makecell[c]{\textbf{6.26}}         &\makecell[c]{\textbf{52.99}}
			 &\makecell[c]{\textbf{52.84}}\\ 			 

		\hline	 
		\end{tabular}
	\end{center}
	\caption{Ablation study of our GMF on 3DRegNet.}
	\label{t8}
\end{table}
We also report the time information of each component. Table \ref{t1} shows that our module is very efficient.
\begin{table}[h]
	\scriptsize
	\begin{center}
		\begin{tabular}{c|cccc}		
			
			\hline
			\textbf{Component} 
			& {F-2}
			& {F-1}
			& {LCPE} 
			& {PDSC}\\ 
			\hline
			\textbf{time{s}}
			& {$1.7 \times 10^{-3}$ } 
			& {$1.8 \times 10^{-3}$} 
			& {$2.7 \times 10^{-4}$}
			& {0.13}\\ 
			
			\hline
		\end{tabular}
	\end{center}
	\caption{Time for every component on PointDSC framework. F-2 means the fusion layer 2, F-1 means the fusion layer 1, LCPE means the LCPE layer, PDSC means the PointDSC correspondence feature extraction module.}
	\label{t1}
\end{table}

\section{Conclusions}
In this paper, we propose a general multimodal fusion framework for correspondence outlier rejection to improve the point cloud registration accuracy. The proposed framework has two advantages. Firstly, it leverages both texture and structure information to improve the distinctiveness of correspondence feature. Secondly, the convolution position encoding enables the cross-attention operation integrate both local and global information. The experiments results show that the proposed module achieves wide generalization ability and consistently improve the correspondence outlier rejection accuracy and registration accuracy on both indoor and outdoor datasets. The proposed framework is also robust to different light conditions and noise.

{\small
\bibliographystyle{IEEEbib}
\bibliography{egbib}
}

\end{document}